\documentclass{article}

\usepackage{PRIMEarxiv}

\usepackage[T1]{fontenc}    
\usepackage[hidelinks]{hyperref}       
\usepackage{url}            
\usepackage{booktabs}       
\usepackage{amsfonts}       
\usepackage{amssymb,amsthm}
\usepackage{multirow}
\usepackage{nicefrac}       
\usepackage{microtype}      
\usepackage{caption}        
\usepackage{lipsum}
\usepackage{amsmath}
\usepackage{fancyhdr}       
\usepackage{graphicx}       
\usepackage[inkscapelatex=false]{svg}
\graphicspath{{media/}}     

\definecolor{gold}{rgb}{1.0, 0.84, 0.0} 
\definecolor{silver}{rgb}{0.75, 0.75, 0.75} 
\definecolor{bronze}{rgb}{0.8, 0.5, 0.2}

\hypersetup{
  colorlinks  = true, 
  urlcolor    = blue, 
  linkcolor   = blue, 
  citecolor   = red 
}

\title{Unlocking the Hidden Potential of CLIP in Generalizable Deepfake Detection
}

\author{
  Andrii Yermakov, Jan Cech, Jiri Matas \\
  Visual Recognition Group \\
  Faculty of Electrical Engineering \\
  Czech Technical University in Prague \\
  \texttt{\{yermaand, cechj, matas\}@fel.cvut.cz} \\
}

\begin{document}
\maketitle

\begin{abstract}

This paper tackles the challenge of detecting partially manipulated facial deepfakes, which involve subtle alterations to specific facial features while retaining the overall context, posing a greater detection difficulty than fully synthetic faces. We leverage the Contrastive Language-Image Pre-training (CLIP) model, specifically its ViT-L/14 visual encoder, to develop a generalizable detection method that performs robustly across diverse datasets and unknown forgery techniques with minimal modifications to the original model. The proposed approach utilizes parameter-efficient fine-tuning (PEFT) techniques, such as LN-tuning, to adjust a small subset of the model's parameters, preserving CLIP's pre-trained knowledge and reducing overfitting. A tailored preprocessing pipeline optimizes the method for facial images, while regularization strategies, including L2 normalization and metric learning on a hyperspherical manifold, enhance generalization. Trained on the FaceForensics++ dataset and evaluated in a cross-dataset fashion on Celeb-DF-v2, DFDC, FFIW, and others, the proposed method achieves competitive detection accuracy comparable to or outperforming much more complex state-of-the-art techniques. This work highlights the efficacy of CLIP's visual encoder in facial deepfake detection and establishes a simple, powerful baseline for future research, advancing the field of generalizable deepfake detection. The code is available at: \url{https://github.com/yermandy/deepfake-detection}

\end{abstract}


\section{Introduction}
\label{sec:intro}

This work studies \textbf{partially manipulated facial deepfakes}, which are facial images or videos altered through techniques like \textbf{face reenactment} or \textbf{face swapping}, rather than fully generated faces created entirely from scratch. Unlike fully synthetic faces, these partial manipulations involve modifying specific aspects -- such as expressions or identities -- while preserving much of the original context, making them harder to detect due to their subtle localized artifacts. Our research focuses on developing a generalizable detection method tailored to these types of deepfakes, leveraging the power of \textbf{Contrastive Language-Image Pre-training} (CLIP)~\cite{CLIP}, a pre-trained model known for its rich, generalizable representations. We will show that by introducing a few tricks to the pre-trained CLIP, the proposed approach can achieve results similar to the current state-of-the-art methods.

In this work, we harness CLIP's visual encoder, specifically the ViT-L/14 variant, to tackle \textbf{facial deepfake detection}. We developed a detection method that performs strongly across multiple datasets and forgery techniques with minimal modifications to the original model. To this end, we employ \textbf{parameter-efficient fine-tuning} (PEFT)~\cite{PEFT} techniques, such as LN-tuning~\cite{LN-tuning}, which adjusts only a small subset of the model's parameters. This approach preserves CLIP's pre-trained knowledge while reducing computational costs and mitigating overfitting -- a common challenge in deepfake detection due to limited data availability.

A custom preprocessing pipeline, consisting of face detection, alignment, and cropping, is employed to optimize the proposed approach for \textbf{facial images}~\cite{DeepfakeBench}. This helps the model to focus on the most relevant features for detecting facial deepfakes. Additionally, we enhance generalization through regularization strategies, such as \textbf{L2 normalization}, which projects features onto a hypersphere. To improve the separation between real and fake classes, we further explore metric learning on this hyperspherical manifold~\cite{UniAlign} to promote tighter clustering within classes and greater distinction between them, strengthening the model's robustness to unseen forgery methods and datasets.

We evaluate the proposed approach on benchmarks such as \textbf{FaceForensics++}~\cite{FF++}, \textbf{Celeb-DF-v2}~\cite{CDFv2}, \textbf{Google's DFD}, \textbf{DeepFake Detection Challenge}~\cite{DFDC}, \textbf{Face Forensics in the Wild}~\cite{FFIW}, and \textbf{DeepSpeak v1.0}~\cite{DSv1} which include face swapping and reenactment forgeries. The results show that the proposed method performs competitively with state-of-the-art techniques, effectively detecting partially manipulated facial deepfakes with minimal modifications to CLIP. This work provides a strong baseline for identifying these subtle manipulations, distinct from fully generated faces, and contributes to advancing deepfake detection research.

Our key contributions are:

\begin{itemize}
    \item We showcase the effectiveness of CLIP's visual encoder adjusted for facial deepfake detection, creating an easy-to-implement pipeline that achieves results close to the current state-of-the-art methods.

    \item We present comprehensive experimental results across multiple benchmarks, confirming the robustness and adaptability of the proposed approach to facial deepfake detection.
\end{itemize}

\section{Related Works}
\label{sec:related_works}

The facial deepfake detection field has seen significant advancements, with various methods addressing the challenge of detecting manipulated facial content across diverse datasets and forgery techniques. A central challenge in this domain is preparing for samples produced by out-of-distribution (OOD) deepfake generators. With the scarcity of deepfake video datasets and dissimilar generator fingerprints (cues left in samples by the generation process), generalization to samples from newly appearing generators becomes a primary issue. Furthermore, as generation pipelines evolve, they leave fewer detectable traces, diminishing the effectiveness of detection methods that depend on specific inductive biases, such as inconsistencies in blending boundaries~\cite{X-ray}, head poses~\cite{inconsistent-head}, eye blinking patterns~\cite{inconsistent-blinking}, or lip movements~\cite{LipForensics}. In our approach, we address this challenge by focusing solely on the presence of subtle generation fingerprints within the extracted facial image, leveraging the discriminative features of CLIP's visual encoder, which are further refined through a set of simple techniques detailed in Section~\ref{sec:method}.

Our work draws inspiration from~\cite{LSDA}, which augments the latent space to enhance generalization and introduces a novel augmentation strategy on the hyperspherical manifold, elaborated in Section~\ref{sec:slerp}.

Recent efforts in detecting general AI-generated content~\cite{UniFD, CLIPping, SSL-in-DF, FatFormer} and facial deepfakes specifically~\cite{UDD, ForensicsAdapter, Effort} have begun to adopt CLIP as a backbone for generalizable deepfake detection. The most recent papers in facial deepfake detection appeared on arXiv while we conducted our experiments. Even though there are some similarities between recent works, we conducted our research independently, achieving similar or better results with a simpler strategy, setting an even simpler baseline for future comparisons. Additionally, we provide open access to our code and model weights to facilitate reproducibility~\footnote{\url{https://github.com/yermandy/deepfake-detection}}.

\section{Method}
\label{sec:method}

\subsection{Baseline model}

In this paper, we study the effectiveness of Contrastive Language-Image Pre-training (CLIP)~\cite{CLIP} in facial deepfake detection. CLIP features, known for their ability to generalize across new datasets and generative models, as well as their resilience to post-processing, have demonstrated significantly superior performance in the identification of AI-generated content compared to other popular large-scale foundation models~\cite{UniFD, CLIPping, LipFD, FatFormer}.

As a baseline model, we use the CLIP-ViT-L/14 image encoder trained by OpenAI~\cite{CLIP}. This model has $16\times16 + 1$ tokens of dimension 1024. $16 \times 16$ represents the number of visual patches, and 1 refers to the classification token. We use only the classification token and discard all patch tokens. We freeze all weights to preserve the learned representations and train a single linear binary classifier with the cross-entropy loss function using the classification token as the input. In total, we train only weights $W\in\mathbb{R}^{1024\times2}$ and biases $b\in\mathbb{R}^2$. To our knowledge, this baseline in the context of detection of AI-generated content was first introduced in~\cite{UniFD} paper.

\subsection{Dataset preprocessing}

We found that while working with facial images, it is crucial to perform standardized facial processing for training and validation splits. We assume that when face swapping or face reenactment is applied to a facial image, the detector expects to see forgery cues in certain regions, which improves performance and makes the task easier to solve. Also, for face reenactment, where only a small area is changed (e.g., lip movements), the resolution plays an important role. For the standardized processing, we employ the pipeline proposed in DeepfakeBench~\cite{DeepfakeBench}, see further details in Section~\ref{sec:preprocessing}.

\subsection{Parameter-efficient fine-tuning}

Similarly to~\cite{SSL-in-DF, CLIPping, FatFormer}, we explore parameter-efficient fine-tuning (PEFT) strategies, which fine-tune only a small (compared to the total) number of parameters, while keeping most parameters untouched, thus efficiently introducing several degrees of freedom. PEFT is better for generalization than full fine-tuning in low data regimes~\cite{PEFT}, which is particularly applicable to deepfake detection, where collected datasets and benchmarks have a low number of samples. This work explores three PEFT methods: LoRA~\cite{hu2022lora} and LN-tuning~\cite{LN-tuning, CLIPFit} as bias tuning~\cite{CLIPFit}. LN-tuning is done the same way as in~\cite{CLIPFit}. Regarding bias tuning, unlike~\cite{CLIPFit}, we focus exclusively on the vision encoder of CLIP, and therefore, we adjust the biases within the MLP layers of the vision encoder.

\subsection{Regularization of the feature manifold to a hypersphere}
\label{sec:reg-of-feature-manifold}

We observed that restricting the features (outputs of the penultimate layer, before the classification layer) to unit length by L2 normalization can improve the generalization to unseen datasets and forgery methods. This normalization can be viewed as a projection from Euclidean space to the hypersphere manifold. It was empirically observed that restricting features to live in a hypersphere helps to achieve better results in face-related tasks such as face recognition~\cite{schroff2015facenet, deng2019arcface}. From the perspective of a linear classifier, classification is akin to separating a spherical cap from the rest of a hypersphere. This encourages the model to create clusters on a hyperspherical cap and serves as a type of regularization.

\subsection{Metric learning on a hypersphere}

Building on the concept of feature manifold regularization, we aim to impose further constraints by introducing additional priors. One potential prior is to bring features of the same class closer, creating denser clusters and increasing the distance between dissimilar classes. Such a prior can further improve generalization. In this work, we experiment with the supervised contrastive loss~\cite{SupCon} as well as directly optimizing the uniformity and alignment metrics~\cite{UniAlign}.

\subsection{Latent augmentations}
\label{sec:slerp}

Similarly to~\cite{LSDA}, we experiment with adding augmentations to extracted features. As we constrain features to live in a hyperspherical manifold, we take another approach to augmenting the latent space. We use spherical linear interpolation (slerp) between features from the same class in the same batch. For each feature $x_i$ in a batch, we randomly sample a paired feature $x_j$ having the same class $y_i=y_j$ and sample an interpolation parameter $t\sim\mathcal{U}(0,1)$ from the uniform distribution. The slerp can then be defined as:

\begin{equation}
\text{slerp}(\mathbf{x}_i, \mathbf{x}_j; t) = 
\frac{\sin((1 - t) \theta)}{\sin \theta} \mathbf{x}_i + 
\frac{\sin(t \theta)}{\sin \theta} \mathbf{x}_j
\label{eq:slerp}
\end{equation}

where $\theta = \arccos (\mathbf{x}_i^{\intercal} \mathbf{x}_j)$, with $\|\mathbf{x}_i\| = \|\mathbf{x}_j\| = 1.$ For the fake class, this augmentation helps to fill the latent space between feature clusters formed by samples from known generators. We believe this can potentially prepare the detector for newly appearing generators, forgery fingerprints of which can resemble some combination of features from the known generators. This augmentation works as another form of regularization.

\section{Experiments}
\label{sec:experiments}

\subsection{Datasets}

For training and comparison purposes, we are using the FaceForensics++ (FF++) dataset~\cite{FF++}, which consists of 3600 videos, from which 720 are real videos and the rest $4\times720$ come from four different deepfake forgery methods, see Table~\ref{tab:train_dataset_statistics}. The dataset was released in three different compression levels. We employ the c23 version to facilitate a fair comparison with alternative methods. 

We noticed that the usage of the FF++ validation set does not resemble the distribution of the test set and is very similar to the training set. The comparison of methods used in the ablation studies is not possible in these settings, as each of them achieves the best scores quickly and does not show the overfitting effects. The FF++ validation set does not provide crucial information on how methods perform during longer training. Because of this, we create our own validation set consisting of a subset (15\%) of the test set as well as samples from other deepfake datasets and with deepfakes created by us using FaceFusion software~\footnote{\url{https://github.com/facefusion/facefusion}}. In our experiments, we will show that using this validation set, we can find the set of regularization techniques that allows us to select most checkpoints without compromising on performance on the validation set.

For testing, we are using Celeb-DF-v2 (CDFv2)~\cite{CDFv2}, DeepFake Detection Challenge (DFDC)~\cite{DFDC}, Google's DFD dataset~\footnote{\url{https://research.google/blog/contributing-data-to-deepfake-detection-research/}}, Face Forensics in the Wild (FFIW)~\cite{FFIW}, DeepSpeak v1.0 (DSv1)~\cite{DSv1}. Dataset statistics are shown in Table~\ref{tab:test_dataset_statistics}.

\begin{table}[ht]
\begin{minipage}{0.45\textwidth}
    \centering
    \caption{Train dataset statistics. All models were trained on the FF++~\cite{FF++} dataset. The training split consists of 3600 videos, from which we sampled 22352 frames. There are 720 videos of the real class and $4\times720$ videos belonging to four deepfake methods.}
    \label{tab:train_dataset_statistics}
    \begin{tabular}{cccc}
        \toprule
        \textbf{Dataset} & \textbf{Source} & \textbf{Videos} & \textbf{Frames} \\
        \midrule
        \multirow{5}{*}{FF++~\cite{FF++}}
        & Real & 720 & 4479\\
        & DF & 720 &   4437  \\
        & F2F & 720 &  4480 \\
        & FS & 720 &   4477 \\
        & FN & 720 &   4479  \\
        \bottomrule
    \end{tabular}
\end{minipage}
\hfill
\begin{minipage}{0.50\textwidth}
    \centering
    \captionof{table}{Test dataset statistics. Values represent the number of videos in real and fake classes used in our experiments. Values in parentheses represent the number of videos missing from our test set compared to the original datasets due to the failure of the DLIB~\cite{DLIB} object detector.}
    \label{tab:test_dataset_statistics}
    \begin{tabular}{lcc}
        \toprule
        \textbf{Dataset} & \textbf{Real} & \textbf{Fake} \\
        \midrule
        DFD & 363 & 3066 (+2) \\
        CDFv2~\cite{CDFv2} & 178 & 340 \\
        DFDC~\cite{DFDC} & 2315 (+185) & 2389 (+111) \\
        FFIW~\cite{FFIW} & 1705 (+33) & 1705 (+33) \\
        DSv1~\cite{DSv1} & 1324 (+1) & 1494 (+6) \\
        \bottomrule
    \end{tabular}
\end{minipage}
\end{table}

\subsection{Preprocessing}
\label{sec:preprocessing}

We use the preprocessing pipeline proposed in DeepfakeBench~\cite{DeepfakeBench} using their code~\footnote{\url{https://github.com/SCLBD/DeepfakeBench}}. The pipeline is:

\begin{enumerate}
  \item Samples 32 frames from each video evenly.
  \item Extracts the biggest face using the DLIB~\cite{DLIB} detector.
  \item Aligns the face using predicted landmarks.
  \item Calculates the bounding box for the aligned face.
  \item Expands the bounding box by a $1.3\times$ margin.
  \item Crops and resizes the face to a $256\times256$ RGB image
  \item Saves the image in PNG format.
\end{enumerate}

Notice that the actual input to the CLIP-ViT-L/14 model is not $256\times256$ but $244\times244$. The CLIP-ViT-L/14 preprocessor handles this further resizing, together with image normalization, to get the expected input.

\subsection{Metrics}
\label{sec:metrics}

We use the video-level area under the ROC curve (AUROC) as our main comparison and optimization metric. To get a score value for a video, we take the mean value across 32 predictions from every sampled frame of this video, see~\ref{sec:preprocessing}. Precisely, for AUROC computation, we are using the one-vs-rest strategy with macro averaging, which corresponds to \verb|roc_auc_score(y_true, y_score, multi_class="ovr", average="macro")| in \verb|sklearn| implementation. All values in the tables are referenced to this metric unless specified differently.

\subsection{Training setup}

In our experiments, we use the visual part of CLIP-ViT-L/14 as the main backbone. To optimize model parameters, we use the Adam~\cite{Adam} optimizer without weight decay. Betas are set to $\beta_1=0.9$ and $\beta_2=0.999$, weight precision is bfloat16, and the learning rate is scheduled with a cosine decay. The initial learning rate is 8e-5 and decays to 5e-5 after 50 epochs of training, even though most runs stop improving after 15 epochs. The batch size is set to 128 samples.

\subsection{Image augmentations}

We augment the training set using augmentations such as random horizontal flip, random affine transformation, Gaussian blurring, color jitter, and JPEG compression implemented in the torchvision package.

\section{Results}

\begin{table}[ht]
\centering
\caption{Generalization of models trained on the FF++ dataset to unseen datasets and forgery methods. Reported values are \textbf{video-level AUROC} as described in Section~\ref{sec:metrics}. Results of other methods are taken from their original papers. Values with superscript citations are taken from the papers referenced in superscripts.}
\label{tab:methods-comparison}
\begin{tabular}{lccccccc}
\toprule
\textbf{Model} & \textbf{Year} & \textbf{Publication} & \textbf{CDFv2} & \textbf{DFD} & \textbf{DFDC}  & \textbf{FFIW}  & \textbf{DSv1}\\
\midrule
LipForensics~\cite{LipForensics}    & 2021 & CVPR  & 82.4           & --             & 73.5  &   --    &--\\
FTCN~\cite{FTCN}                    & 2021 & ICCV  & 86.9           & --             & 74.0  & $74.47^\text{\cite{SBI}}$         &--\\
RealForensics~\cite{RealForensics}  & 2022 & CVPR  & 86.9           & --             & 75.9  &   --      &--\\
SBI~\cite{SBI}                      & 2022 & CVPR  & 93.18          & 82.68          & 72.42 &  84.83  &--\\
AUNet~\cite{AUNet}                  & 2023 & CVPR  & 92.77          & 99.22          & 73.82 &  81.45  &--\\
StyleDFD~\cite{StyleDFD}            & 2024 & CVPR  & 89.0           & 96.1           & --    &   --      &--\\
LSDA~\cite{LSDA}                    & 2024 & CVPR  & 91.1           & --             & 77.0  & $72.4^{\text{\cite{Effort}}}$  &--\\
LAA-Net~\cite{LAA-Net}              & 2024 & CVPR  & 95.4           & 98.43          & 86.94 &   --      &--\\
AltFreezing~\cite{AltFreezing}      & 2024 & CVPR  & 89.5           & 98.5  & 99.4  &   --      &--\\
NACO~\cite{NACO}                    & 2024 & ECCV  & 89.5           & --             & 76.7  &   --      &--\\
TALL++~\cite{TALL++}                & 2024 & IJCV  & 91.96          & --             & 78.51 &   --      &--\\
UDD~\cite{UDD}                      & 2025 &  arXiv   & 93.13          & 95.51          & 81.21 &   --      &--\\
Effort~\cite{Effort}                & 2025 &  arXiv   & 95.6           & 96.5           & 84.3  & 92.1  &--\\
KID~\cite{KID}                      & 2025 &  arXiv   & 95.74           & 99.46           & 75.77  & 82.53  &--\\
ForensicsAdapter~\cite{ForensicsAdapter}  & 2025 &  arXiv   & 95.7     & 97.2           & 87.2  & --  &--\\
\midrule
\textbf{Proposed}                       & 2025 &  arXiv   & 96.62 & 98.0          & 87.15 & 91.52  &92.01\\
\bottomrule
\end{tabular}
\end{table}

\subsection{Quantitative Analysis}

Table~\ref{tab:methods-comparison} compares the video-level AUROC of the proposed method with state-of-the-art approaches. The proposed method achieves AUROC scores of 96.62 on CDFv2, 98.0 on DFD, 87.15 on DFDC, 91.52 on FFIW, 92.01 on DSv1 demonstrating a compelling generalization across datasets while requiring minimal changes to the original CLIP parameters. 

\subsection{Ablation studies}

To understand the contribution of each component, we conducted ablation experiments. They include:

\begin{enumerate}
    \item \textbf{Linear Probing} -- the setup is similar to~\cite{UniFD}. It only adds a linear classifier on top of features from the CLIP ViT classification token.
    \item \textbf{LN-Tuning} besides training the linear classifier, unfreezes linear norm layers. It finetunes only 104K out of 303M total model parameters, which is roughly only 0.03\% of all model parameters.
    \item \textbf{LN-Tuning + Norm} -- adds L2 normalization of features to the previous setup
    \item \textbf{LN-Tuning + Norm + UnAl} -- adds uniformity and alignment loss to the previous setup
    \item \textbf{LN-Tuning + Norm + UnAl + Slerp} -- adds slerp augmentations to the previous setup
\end{enumerate}

\begin{figure}[ht]
    \centering
    \includegraphics[width=0.55\linewidth]{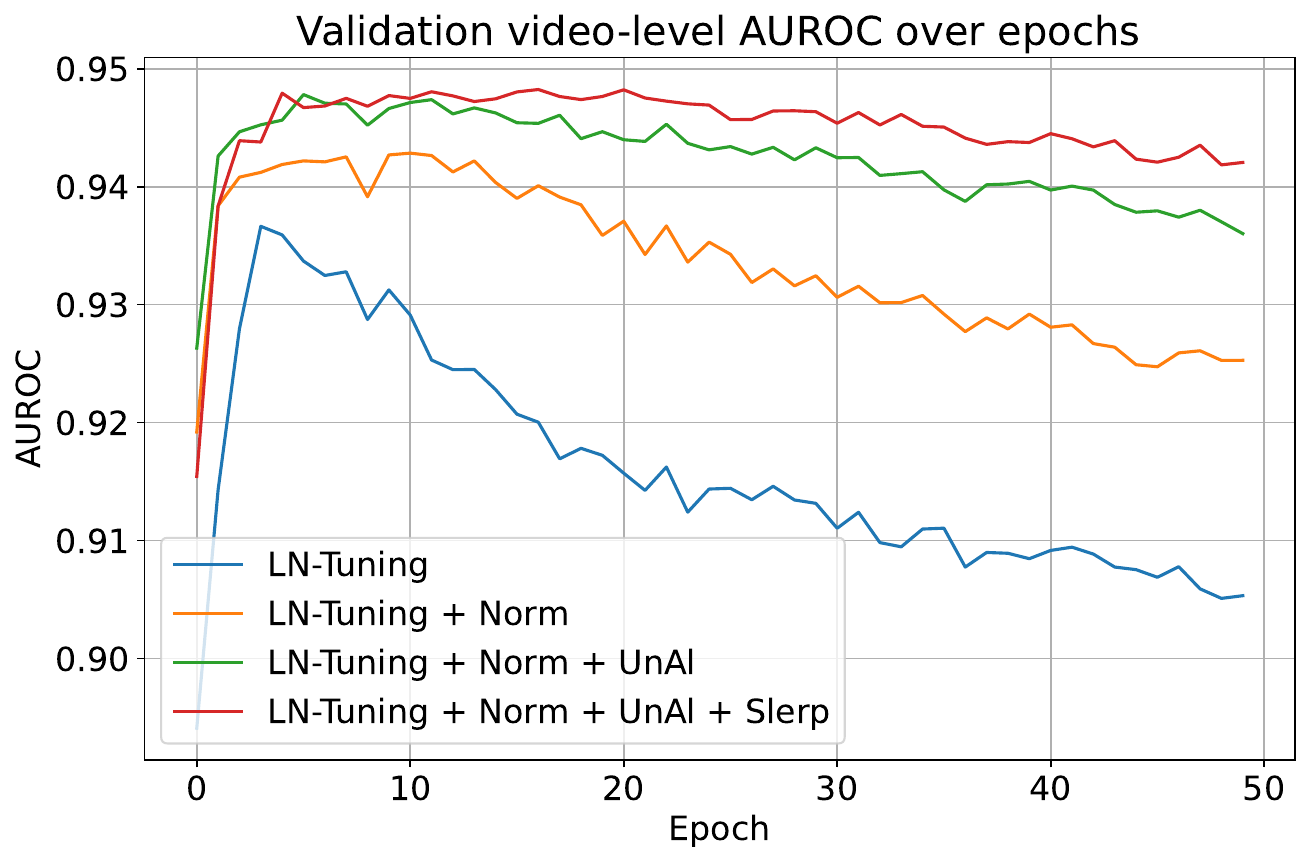}
    \caption{Impact of different regularization methods on validation video-level AUROC. \textbf{LN-Tuning} unfreezes layer norm layers. \textbf{Norm} incorporates the L2 normalization of features before the classification layer. \textbf{UnAl} incorporates uniformity and alignment losses. \textbf{Slerp} incorporates spherical linear interpolation between L2 normalized features.}
    \label{fig:overfittung}
\end{figure}

\begin{table}[ht]
    \centering
    \begin{tabular}{lccccc}
        \toprule
        \textbf{Setup} & \textbf{CDFv2} & \textbf{DFD} & \textbf{DFDC} & \textbf{FFIW} & \textbf{DSv1} \\
        \midrule
        (1)\; Linear Probing & 78.13 & 88.15 & 73.62 & 79.28 & 63.65 \\
        (2)\; LN-Tuning & 94.88 & 96.83 & 86.41 & 92.24 & 83.57 \\
        (3)\; LN-Tuning + Norm & 96.21 & \textbf{98.18} & \textbf{87.82} & \textbf{92.72} & 88.81 \\
        (4)\; LN-Tuning + Norm + UnAl & \textbf{96.64} & 97.96 & 86.97 & 91.76 & 91.16 \\
        (5)\; LN-Tuning + Norm + UnAl + Slerp & 96.62 & 98.00 & 87.15 & 91.52 & \textbf{92.01} \\
        \bottomrule
    \end{tabular}
    \caption{Impact of regularization methods on test video-level AUROC}
    \label{tab:regularizations-comparison}
\end{table}

Figure~\ref{fig:overfittung} shows the impact of each added component on validation AUROC across training epochs, confirming that each component contributes to the overall model regularization. Additionally, we present the test AUROC results in Table~\ref{tab:regularizations-comparison} to show what influence each component has on an out-of-distribution dataset and forgery method.

\paragraph{PEFT} Our experiments showed that LN-tuning achieves the best performance and prevents overfitting compared to other PEFT strategies such as bias tuning or LoRA. We noticed that setting the LoRA rank to one, which is the lowest possible, allows the model to achieve perfect training AUROC (99.99\%) in one training epoch, leading to quick overfitting.

\paragraph{Feature Normalization} Incorporation of L2 normalization results in a better validation AUROC and its slower decrease during longer training, indicating its importance in mitigating overfitting. This single change is the most important among all tried regularizations leading to the biggest and most pronounced improvements, see rows (2) and (3) in Table~\ref{tab:regularizations-comparison}.

\paragraph{Metric Learning} We did not achieve as good results with supervised contrastive loss as using uniformity and alignment. It might be explained by the fact that the batch size was not big enough (128 samples), which was a memory constraint for a single NVIDIA A100-SXM4-40GB. The addition of uniformity and alignment loss leads to slower overfitting and better validation AUROC. We noticed that even the addition of uniformity loss makes the results better and close to the addition of both. The weighted combination of the uniformity loss with cross-entropy regularizes the hyperspherical space in such a way that uniformity loss distributes features uniformly across the unit hypersphere, while the cross-entropy makes the clusters tighter. We argue that this can make use of the hyperspherical space more effectively and lead to a better generalization.

\paragraph{Latent Augmentations} We noticed that incorporating latent augmentations such as slerp, described in Section~\ref{sec:slerp}, acts as a form of regularization, effectively preventing overfitting. This augmentation enriches the feature space and can help to make the representation of the fake class more robust to samples from unseen generation techniques, whose representations can be similar to some combinations of previously seen ones.

\section{Conclusion}
\label{sec:conclusions}

In this paper, we explored alternative ways to demonstrate the generalizable power of the visual encoder of the CLIP model in the problem of deepfake detection. We adapted CLIP using parameter-efficient fine-tuning techniques, specifically LN-tuning, to tailor it to the deepfake detection task while minimizing overfitting. Additionally, we explored regularization methods such as feature normalization and metric learning on a hypersphere to constrain the feature space and enhance the model's ability to generalize across unseen datasets and forgery methods.

Our experimental results on widely adopted benchmarks demonstrate comparable performance to more complex state-of-the-art methods. Ablation studies further validate the effectiveness of each component, underscoring the benefits of parameter-efficient fine-tuning and proposed regularization techniques in achieving robust generalization.

Despite these promising outcomes, challenges remain. For example, performance on datasets like DFDC could be further improved, potentially by exploring additional pre-trained models or multimodal approaches. The proposed method focuses on frame-level analysis, leaving room for future work to incorporate spatiotemporal information, as well as sound for video-based detection.

Overall, this research highlights the potential of large-scale pre-trained models like CLIP in addressing the evolving challenge of deepfake detection. The proposed approach provides an easy-to-implement but powerful baseline and a foundation for building more robust and generalizable systems, paving the way for future advancements in this critical domain.

\section*{Acknowledgments}

The research was supported by the CEDMO 2.0 NPO (MPO 60273/24/21300/21000) and CTU Student Grant SGS23/173/OHK3/3T/13.

\bibliographystyle{unsrt}  
\bibliography{references}

\end{document}